\documentclass{article}

\usepackage{arxiv}

\usepackage[utf8]{inputenc} 
\usepackage[T1]{fontenc}    
\usepackage{hyperref}       
\usepackage{url}            
\usepackage{booktabs}       
\usepackage{amsfonts}       
\usepackage{nicefrac}       
\usepackage{microtype}      
\usepackage{lipsum}
\usepackage{fancyhdr}       
\usepackage{graphicx}       
\graphicspath{{media/}}     

\title{Improved Breast Cancer Diagnosis through Transfer Learning on Hematoxylin and Eosin Stained Histology Images}

\author{ \href{https://orcid.org/0000-0001-9201-1580}{\includegraphics[scale=0.06]{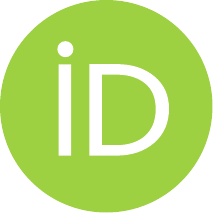}\hspace{1mm}Fahad Ahmed}\\
 TIB - Leibniz Information Center for Science and Technology, Hannover, Germany\\
 CARESAI\\
 \texttt{fahad.ahmed@tib.eu}
 \And
 Reem Abdel-Salam\\
 Cairo University, Faculty of Engineering, Computer Department, Egypt\\
 CARESAI\\
 \texttt{reem.abdelsalam13@gmail.com}
 \And
 Leon Hamnett\\
 CARESAI\\
 \texttt{L.s.hamnett@gmail.com}
 \And
 Mary Adewunmi\\
 College of Health \& Medicine(CHM), School of Medicine, University of Tasmania(UTAS), Hobart, Australia\\
 CARESAI\\
 \texttt{mary.adewunmi@utas.edu.au}
 \And
 Temitope Ayano\\
 Department of Microbiology, Obafemi Awolowo University, Ile-Ife, Nigeria\\
 CARESAI\\
 \texttt{ayanotemitope@gmail.com}
}

\begin{document}
\maketitle
\begin{abstract}
Breast cancer is one of the leading causes of death for women worldwide. Early screening is essential for early identification, but the chance of survival declines as the cancer progresses into advanced stages. For this study, the most recent BRACS dataset of histological (H\&E) stained images was used to classify breast cancer tumours, which contains both the whole-slide images (WSI) and region-of-interest (ROI) images, however, for our study we have considered ROI images. We have experimented using different pre-trained deep learning models, such as Xception, EfficientNet, ResNet50, and InceptionResNet, pre-trained on the ImageNet weights. We pre-processed the BRACS ROI along with image augmentation, upsampling, and dataset split strategies. For the default dataset split, the best results were obtained by ResNet50 achieving 66\% f1-score. For the custom dataset split, the best results were obtained by performing upsampling and image augmentation which results in 96.2\% f1-score. Our second approach also reduced the number of false positive and false negative classifications to less than 3\% for each class. We believe that our study significantly impacts the early diagnosis and identification of breast cancer tumors and their subtypes, especially atypical and malignant tumors, thus improving patient outcomes and reducing patient mortality rates. Overall, this study has primarily focused on identifying seven (7) breast cancer tumor subtypes, and we believe that the experimental models can be fine-tuned further to generalize over previous breast cancer histology datasets as well.
\end{abstract}

\section{Introduction}
Breast cancer is one of the major causes of death among women worldwide \cite{cancer_net,giaquinto2022breast}. Early identification can minimize fatality rates, thereby, preventing disease mutation and progression \cite{ott2009importance}.  A pathologist examine tissue biopsy to confirm the presence of cancerous cells in breast \cite{reshma2022detection}. The tissue is fixed to slides and examined in a lab after the biopsy is finished. Formalin fixation is the first step in the tissue preparation procedure, followed by embedding in paraffin sections \cite{al2021standard}. The paraffin blocks are subsequently cut into pieces and mounted on glass slides. However, the tissue's cytoplasm and nuclei, which contain valuable information for the pathologist, are not yet visible.

Hematoxylin and eosin (H\&E) staining \cite{boschman2022utility}, a common and well-known staining procedure for tissue visibility. Hematoxylin can bind to deoxyribonucleic acid when it is added to the tissue, which causes the cell nuclei to become stained blue or purple. The stroma and cytoplasm, on the other hand, are colored pink because eosin has the ability to bind to proteins and dye these structures. The glass slide is typically coverslipped and set in place after staining and sent to a pathologist for review. The expert routinely collects data on the nuclei's size, shape, organization, interactions, and spatial configurations. The variability, density, and general structure of the tissue are also examined. For identifying carcinoma (a type of cancer that forms in epithelial) and non-carcinoma cells, information on the characteristics of the nuclei, such as shape and density is particularly important. Additionally, knowledge of tissue structure is important for identifying benign, atypical, and malignant classes of cells.

In this study, we used the up-to-date BRACS dataset \cite{brancati2022bracs} to develop our image classification models of histopathological (H\&E) stained images. The ductal carcinoma in situ (DCIS) and invasive carcinoma (IC) lesions are malignant tissues that need immediate medical attention because if left untreated, they will continue to progress and will cause a fatality in the patient. In-situ carcinoma refers specifically to the development of atypical cells that are restricted to the layer of breast tissue from which they originated. Atypical cells that have invaded normal tissue outside of the glands or ducts from which they originated are referred to as invasive carcinoma.

Treatment for invasive carcinoma is challenging because the cancer cells may spread to other organs \cite{akram2017awareness} and the treatment can put the entire body at risk. As the cancer progresses through more advanced stages, the likelihood of survival decreases. Additionally, a patient's in-situ carcinoma tissue may transform into invasive carcinoma tissue in the absence of appropriate and effective treatment \cite{sirinian2022cellular}. Therefore, it is crucial to effectively examine the biopsy tissue so that a diagnosis can be made and treatment can start. It is not an easy or simple task to conduct a practical examination of the tissue. Instead, it takes a lot of time and is, more importantly, prone to human error. Professionals have a diagnostic accuracy rate of about 75\% on average \cite{elmore2015diagnostic}. Patients with incorrect diagnoses could suffer serious and even fatal consequences as a result of these problems.\

The need for a system that can automatically and accurately classify breast cancer histopathology images is generally justified by the lack of trained medical specialists \cite{kandel2020deeply} based on the population of patients they must serve, as well as the laborious process of arriving at a final diagnosis, and the problem of interobserver variability.

While providing adequate classification accuracies based on data availability, previous research solutions have been computationally expensive.  Convolutional neural networks (CNNs) have used multiple instance learning (MIL) to classify cancer histopathology images. The WSI is divided into patches for gigapixel classification. MIL patches are instances in bags. Multiple instances learning is called that because each image (WSI) has many patches, the bag contains several instances representing different patches. Training supervision requires only global picture labels. To make a prediction, the aggregation process summarises all instance data. These multiple instance learning (MIL) models are based on vision transformer models, which are computationally expensive and require large image datasets for effective training.

Patch-based image classification is difficult to apply to high-resolution breast cancer histopathology. Due to large image sizes and high detail, WSIs require a different approach to gain insights. Patch selection approaches are needed because WSIs are divided into patches and cannot fit into memory when training a machine-learning model. If class labels are only available on a WSI basis, it may be difficult to determine which patches contain critical information. No global representative field exists, making model decisions hard to understand and trust.

Due to the above reasons, image classification models that use less memory and require smaller image datasets are needed. Lightweight classification models that perform like current deep learning approaches are essential. This study poses the following main research question: ``Can tuned Exception, EfficientNet, ResNet50, ConvNextTiny, and Inception-ResNet achieve similar results to current state-of-the-art MIL approaches for classifying gigapixel images of breast cancer histology WSIs?'' . Additionally, we explore related questions such as: What pre-trained image classification models are suitable for this task? Can normal patch-based transfer learning work? What data augmentation approaches could improve breast cancer histology slide classification?

This study classifies breast cancer histology gigapixel images using pre-trained image classification models like Xception, EfficientNet, Resnet50, and InceptionResNet, as well as pre-processing techniques like resizes, tiling, stain normalisation, and image compression to improve classification performance. We also use transfer learning to improve classification performance by adapting models from large image datasets to our task. To make models more generalizable and robust to image dataset characteristics, we emphasize augmentation techniques like flipping rotation.

The rest of the paper is organized as follows:
In section 2, Literature Review describes previous related research work and successful strategies. Section 3, Methodology and Techniques, describes this study's framework. Section 4, Discussion and Results details the research and experimentation findings. Section 5 summarizes the paper's findings.

\section{Literature Review}
Breast cancer is a significant public health concern globally, with an estimated 2.3 million new cases identified in 2020 alone \cite{arnold2022current}. Early detection of breast cancer is vital for improved patient outcomes, as treatment success often depends on the stage at which the cancer is diagnosed. Although mammography is the most widely used screening method for breast cancer detection, it has some limitations, particularly in women with dense breast tissue \cite{gordon2022impact}. These limitations include a relatively high false-positive rate, limited sensitivity in dense breast tissue, and the potential risk of radiation exposure due to repeated screening examinations \cite{gordon2022impact}.

Histopathology examination of tissue samples remains the gold standard for breast cancer diagnosis, as it analyzes the image at the cellular level, which is necessary to determine the malignancy of the tumor \cite{belsare2012histopathological}. However, this method requires time-consuming, labor-intensive procedures subject to inter-observer variability. In recent years, deep learning algorithms have shown promising results in automating medical imaging analysis, including histology images, for breast cancer diagnosis.The literature review aims to provide an overview of the current state-of-the-art in deep learning-based approaches for breast cancer diagnosis using histology images.

There has been a growing interest in developing computer-aided diagnosis (CAD) systems for breast cancer using histology images. To address the limitations of previous approaches for multi-classification of histology images, \cite{joseph2022improved} explored the use of handcrafted feature extraction techniques, including Hu moments, Haralick textures, color histograms, and a deep neural network (DNN), for breast cancer multi-classification on the BreakHis dataset. This method performed better in the multi-class classification of breast cancer than most of the related works in the literature. However, \cite{bardou2018classification} compared two machine-learning methods for automatically classifying breast cancer histology images into benign and malignant subclasses. The authors tested dataset augmentation techniques to improve the convolutional neural network's accuracy and found that it outperformed the handcrafted feature-based classifier, achieving accuracy between 96.15\% and 98.33\% for binary classification and 83.31\% and 88.23\% for multi-class classification. To address class imbalance in classifying mitotic and non-mitotic nuclei in breast cancer histopathology images.

\cite{wahab2017two} proposed a CNN architecture and data balancing technique that yielded an F-measure of 0.79, outperforming all methods relying on handcrafted features and those using a combination of handcrafted and CNN-generated features. The results from these studies suggest that CNN-based approaches outperform those relying on handcrafted features, with dataset augmentation techniques and data balancing techniques showing potential for further improving classification accuracy.

Several studies have proposed CNN-based approaches for breast cancer histology image classification. \cite{araujo2017classification} proposed a method that can extract information from relevant scales, including nuclei and overall tissue organization, achieving an overall sensitivity of approximately 81\% for carcinoma patch-wise classification. This proposed system can also be extended to whole-slide breast histology image classification relevant to clinical settings.

\cite{bejnordi2017deep} also proposed a CNN-based CAD system that assesses tumor-associated stroma in hematoxylin and eosin-stained breast specimens. Their method achieves high accuracy for the binary classification of breast tissue biopsies. A class structure-based deep convolutional neural network (CSDCNN) model with multiple hidden layers that achieved high accuracy by preserving intra-class variance using a novel distance constraint for feature space was developed by \cite{han2017breast}, and a capsule network with dynamic routing to classify breast cancer tissue images that achieved promising results for different classes of breast cancer tissue images has also been proposed by \cite{iesmantas2018convolutional}.\par

Furthermore, several studies have proposed novel deep CNN architectures for breast cancer histology image classification. \cite{wei2017deep} proposed a novel deep CNN and advanced data augmentation methods, achieving higher classification accuracy (up to 97\%) with good robustness and generalization. \cite{jiang2019breast} proposed a novel CNN with a small SE-ResNet module and a new learning rate scheduler for automatic classification of histology images into benign and malignant and eight subtypes with high accuracy.

\cite{brancati2018multi} proposed a fine-tuning strategy for ResNet for the multi-classification of breast cancer histology images into normal tissue, benign lesions, in-situ carcinomas, and invasive carcinomas, performing remarkably well on images provided for the Grand Challenge on Breast Cancer Histology Images (BACH) \cite{aresta2019bach}. \cite{yan2020breast} proposed a hybrid convolutional and recurrent deep neural network for breast cancer histopathological image classification. Their model achieved an impressive 95.75\% classification accuracy.

\cite{nazeri2018two} utilized a patch-based deep learning technique consisting of two consecutive convolutional neural networks to achieve 95\% accuracy on the validation set. \cite{rakhlin2018deep} proposed a computational approach based on deep convolutional neural networks for breast cancer histology image classification, achieving an accuracy of 87.2\% and 93.8\% for the 4-class and 2-class classification tasks, respectively.

\cite{tougaccar2020breastnet} proposed a residual architecture model with attention modules and augmentation techniques called BreastNet, achieving stable and accurate classification using the hypercolumn technique, achieving a 98.80\% classification. \cite{nawaz2018classification} modified the convolutional and fully connected layers of a pre-trained AlexNet, achieving patch and image-wise accuracy of 75.73\% and 81.25\%, respectively, and image-wise accuracy of 57\% on the ICIAR-2018 breast cancer challenge hidden test set.

In a study by \cite{mehra2018breast}, transfer learning and fully trained networks for the multi-classification of breast cancer were compared. They found that a fine-tuned, pre-trained VGG16 with a logistic regression classifier had the best performance, achieving 92.60\% accuracy, a 95.65\% area under the ROC curve (AUC), and a 95.95\% accuracy precision score (APS) \cite{mehra2018breast}. \cite{vang2018deep} utilized Inception V3 for patch-level classification and a Dual Path Network (DPN) as a feature extractor for multi-classification of histological images. Their approach achieved a 12.5\% improvement over the state-of-the-art model in experimental results.

Finally, \cite{gour2020residual} designed a 152-layered CNN named ResHist to classify histopathological images into benign and malignant classes using their data augmentation technique. High-throughput Adaptive Sampling for Whole-Slide Histopathology Image Analysis (HASHI) was proposed by \cite{cruz2018high} to detect invasive breast cancer on whole-slide histopathology images (WSI). HASHI uses a CNN-based robust representation learning classifier and a new adaptive sampling method based on probability gradients and quasi-Monte Carlo sampling. Transfer learning has improved breast cancer diagnosis speed and efficiency. Researchers have used pre-trained deep convolutional neural networks (CNNs) to improve breast cancer image classification performance as this enables models to transfer knowledge from domains with substantial image data to domains with limited data.

\section{Methodology and Techniques}
In this research study, we focus on two types of experiments. The first type of experiment involves applying transfer learning on the reduced-resolution images since the original images are extremely high-resolution and also investigating the impact of transfer learning on high-resolution images. In the second type of experiment, we explore the application of transfer learning using image tiling applied to lower-quality and higher-quality images. The dataset used for this project, its statistics, the pre-processing steps taken, the data augmentation techniques, and the training models are all discussed in this section.

\subsection{Dataset}
BRACS \cite{brancati2022bracs} is a set of histopathological slides with hematoxylin and eosin(H\&E) staining applied. It contains atypical lesions as well as six different subtypes of lesions. This also contains histological normal tissue samples. The BRACS dataset is created to support the development of breast cancer diagnostic methods through the automated analysis of histology images. The dataset was developed through the collaboration of the National Cancer Institute - Scientific Institute for Research, (IRCCS) ``Fondazione G. Pascale'', the Institute for High-Performance Computing and Networking (ICAR) of National Research Council (CNR), and IBM Research – Zurich. The dataset was acquired from patients between 2019 and 2020, by board-certified pathologists of the Department of Pathology at the National Cancer Institute- IRCCS "Fondazione G. Pascale" in Naples (Italy).  The dataset used in this study was made available to the public in 2021 \cite{brancati2022bracs}.

\subsection{Dataset Statistics}
The samples were generated from H\&E-stained breast tissue biopsy slides and were selected based on the diagnostic reports of the patients. The age of the patients ranges from 16 to 86 years, with about 61\% of patients in the range of 40-60 years, and only a small number of patients being aged less than 20 years or above 80 years. Due to the laborious and expensive process of data collection, datasets in the medical domain typically consist of relatively small numbers of images. The dataset includes 4539 Regions of Interest (ROIs) that were extracted from 547 Whole-Slide Images (WSIs). Three lesion types are present within the BRACS dataset: benign, malignant, and atypical. These three lesion types are further subtyped into seven groups: Normal (N), Pathological Benign (PB), Usual Ductal Hyperplasia (UDH), Flat Epithelial Atypia (FEA), Atypical Ductal Hyperplasia (ADH), Ductal Carcinoma in Situ (DCIS), and Invasive Carcinoma (IC). The included ROIs have a wide range of image sizes and incorporate common visual artifacts from the tissue-preparation and staining process, that are typical of the types of patterns that would be seen in a clinical setting or pathological laboratory that makes diagnosis more challenging. A total of 7909 breast cancer histopathology image samples, obtained from 82 patients at four different magnification levels, are included in the dataset. The entire dataset takes up about 52 GB of space. The BRACS dataset is a gigapixel dataset (high resolution) comprised of 4391 ROIs obtained from 325 H\&E stained breast cancer WSI with varied sizes and appearances (at magnification level 0.25 µm/pixel for 40x resolution), which can easily exceed 100,000 by 100,000 pixels. The samples in the dataset are separated into two main groups: benign and malignant, with the benign group including 2480 samples and the malignant group including 5429 samples, respectively.

\subsection{Preprocessing}
For the classification of histology images, preprocessing is essential for the stability of model training and for better convergence\cite{yuan2021hybrid}.  
In order to use BRACS dataset with conventional model training approaches, the resolution of the images must be reduced whilst still aiming to preserve important image features. To work around this issue we explore two paths. The first path: is to decrease the resolution of images by resizing and graying out the noise \cite{solem2012programming},  which reduced storage space approximately to 404 MB. The other solution is to apply tiling and patching to the high-resolution images\cite{wu2020patch}. This involves splitting images into multiple smaller images, this retains the level of detail of the original images. The non-uniformity of BRACS ROI images makes it necessary to resize them to a standard image size as suggested by \cite{hsin2016saliency} as typical classification models are not able to accept a mix of image sizes. In our experiments, we focused on 5 different image dimensions: (300 x 300), (512 x 512), (256 x 256), (512 x 256), and (1024 x 1024).

\subsection{Data augmentation}
Data augmentation is a technique used to increase the amount of image variation in the training set because the dataset is much smaller than what is typically needed to train a deep-learning model correctly. This method helps prevent overfitting, a condition in which a model is overly influenced by the information in the training images and is not able to generalize and correctly classify new unseen images. Eight different types of image augmentations are used as data augmentation techniques for both the experiments relating to training models from initial conditions and the experiments involving transfer learning. These image augmentations can be classified as flip (reversing left-right or up-down orientation), rotating, brightening and darkening, resizing, cropping, blurring and sharpening, distorting, noise, and special (emboss, perspective, cutmix, mixup, etc.). A variety of techniques from the Albumentations and MONAI libraries were analysed to see how they contributed to increasing classification accuracy and improving model performance. The process of analyzing histology images is rotationally invariant, as a pathologist can view a microscopy image from any angle and still be able to successfully examine it. As a result, adding a rotation augmentation to the image shouldn't have a negative impact on classification performance. The rotation augmentation is tailored so that an image is rotated 90*k° in a clockwise direction, where k=0,1,2,3. The processed images as discussed in the previous section also received a random width and height shift (translating an image in a certain direction), zoom range (zooming in or out of an image), and horizontal and vertical flips.

\subsection{Data upsampling}
In essence, BRACS is an imbalanced dataset, as shown in Table \ref{data_dirsitnution_brac}. The potential drawbacks of class imbalance within a training dataset are well documented \cite{johnson2019survey_class_imbalance} and include a possible impact on the effectiveness of the model to distinguish between different classes. If the model is not shown a sufficient number of images corresponding to a certain class, it may not effectively learn image features relating to that class and may fail to correctly classify that type of image. Typically, models developed from unbalanced datasets perform poorly in terms of predicting outcomes for the minority class, leading to higher false-negative rates and lower recall \cite{johnson2019survey_class_imbalance}.

\begin{table}[h]
    \centering
    \caption{Original distribution of classes in BRACS dataset.}
    \label{data_dirsitnution_brac}

    \begin{tabular}{c|c|c|c|c|c|c|c}
        \toprule
        Class & N & DCIS & FEA & PB & IC & UDH & ADH \\
        Samples & 484 & 790 & 756 & 836 & 649 & 517 & 507 \\
        \bottomrule
    \end{tabular}
\end{table}

In order to mitigate the issue of unbalanced classes within the dataset, we applied different upsampling strategies such as batch-balanced sampler, weighted batch sampler, and general upsampling. In a batch-balanced sampler, we make sure that each batch contains an equal number of classes.  In the weighted batch sampler, we give weight to each class according to its inverse probability of occurrence, then we sample images in each batch based on their class weights. Finally, in the general upsampling approach, the dataset is upsampled to approximately 1000 and 2000 samples for each class in order to obtain boost performance when classifying under-represented classes.

\subsection{Modeling strategies}
In this research study, we focus on two types of experiments. The first type of experiment involves applying transfer learning on the original images of high resolution as well as on the low-resolution versions. In the second type of experiment, we explore the application of transfer learning using image tiling applied to both high-resolution and lower-quality images.

\subsubsection{Transfer learning}
We implemented 5 well known deep learning classification models, all of which were pre-trained on the ImageNet dataset. These models include Xception, EfficientNet, ResNet50, ConvNextTinyV2, and InceptionResNet. These pre-trained CNN models are often employed in research tasks due to their significant performance against reference classification datasets such as ImageNet and CIFAR-100. Additionally, these models have shown relatively high performance for other challenging computer vision problems within varied research fields \cite{yadav2019deep}.

First, we apply the image pre-processing methodology to the dataset as discussed in the above section. We have investigated several loss functions such as cross-entropy loss function, focal loss, and cross-entropy with label smoothing with a smoothing value of 0.35. AdamW optimizer was used with a learning rate of 1e-3 and weight decay ranges from 1e-2 to 1e-5. In addition, a cosine annealing learning rate scheduler was used. In addition, we investigated the effects of different sampling strategies such as no sampling, weighted balanced sampler, and batch balanced sampler. In our experiments, we have investigated two model architectures.

\begin{figure}[h]
    \centering
    \includegraphics[width=\textwidth]{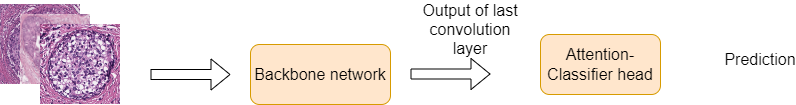}
    \caption{Breast histology image classification workflow by fine-tuning Xception, EfficientNet, ResNet50, ConvNextTinyV2, and InceptionResNet network architectures.}
    \label{arch_1}
\end{figure}

The first architecture is to feed the image directly to the backbone model then the output is fed to the attention classifier layers as shown in figure \ref{arch_1}. In addition, we inspect model performance using two methods original data split and custom data split.

\begin{figure}[h]
    \centering
    \includegraphics[width=\textwidth]{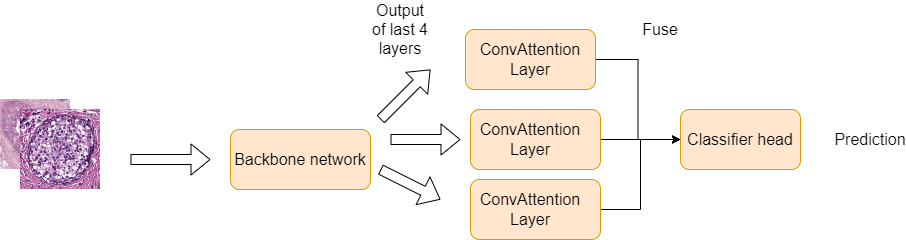}
    \caption{Breast histology image classification workflow by fine-tuning Xception, EfficientNet, ResNet50, ConvNextTinyV2, and InceptionResNet pyramid network architectures.}
    \label{arch_2}
\end{figure}

The second architecture takes inspiration from pyramid networks \cite{hu2020_pyramid_net}. At first, the images are fed to the backbone model. Then we extract the output activation of the last 4 layers. These outputs are then fed to convolution attention layers. Then those outputs are then fused and fed to a classifier head, as shown in figure \ref{arch_2}. Both of these architectures are applied to high-resolution images and low-resolution images.

\subsubsection{Image tiling and transfer learning}
\begin{figure}[h]
    \centering
    \includegraphics[width=\textwidth]{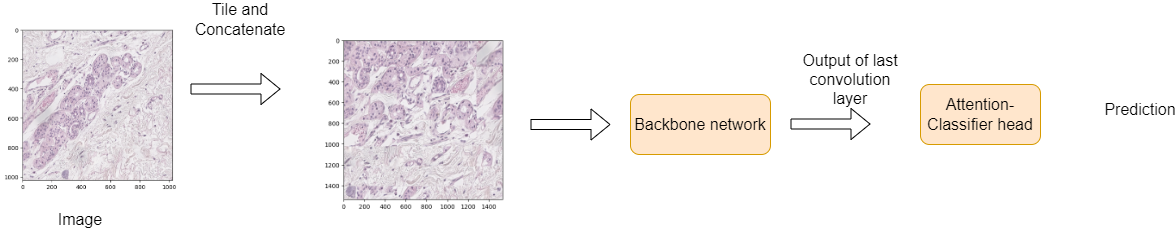}
    \caption{Breast histology image classification workflow by tiling and fine-tuning ResNet50 network.}
    \label{arch_3}
\end{figure}

In this approach, we combine both ideas of multi-instance learning and the standard fine-tuning process. First, for each image, we extract N tiles with different window sizes and zoom levels. Then we concatenate (merge) these tiles along the height or width dimension. This results in a large image size as it contains different tiles and variability. The resultant merged image is of size (1748,1748). The pipeline is shown in figure \ref{arch_3}. In this experiment we have set the image size to (1024, 1024) and the number of tiles to 50, with windows of size 128,256,512 and 3 zoom levels.

\section{Discussion and Results}
We have used weighted f1-score, sensitivity, and accuracy as our main metrics to report performance on the BRACS test-set. In table~\ref{trails_low_resulution_image}, we compare the classification performance of different experiments when making predictions on the test set provided in the BRCAS dataset, using the low-resolution images. The ResNet50 model with focal loss function and image size of (512,512) was able to achieve a high f1-score of 65. The ResNet50 pyramid architecture and the ResNet50 with cutmix and mixup augmentation  both achieve the second-highest classification performance with  an f1-score of 63. These metrics indicate that cutmix and mixup image augmentations, feature pyramids, and a focal loss function are key successes of high performance and training stability. In contrast to other models, ResNet50 showed promising results. Surprisingly, tiling-based methods didn't show high performance, this may need more investigation thoroughly to determine factors such as the best number of tiles to generate per image, as well as tile size, and whether tiles should contain overlapping areas.

\begin{figure}[h!]
    \centering
    \includegraphics[width=7cm, height=7cm]{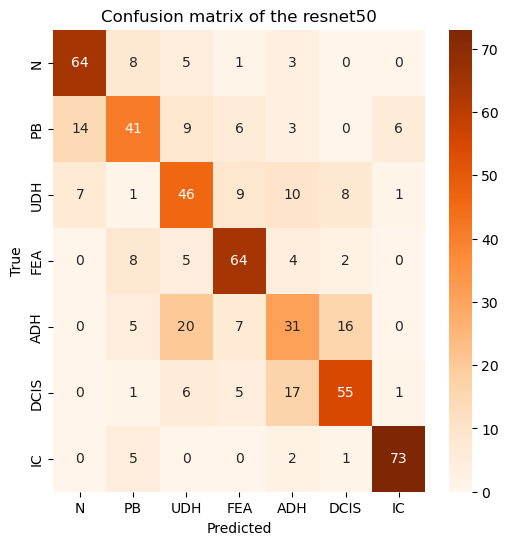}
    \caption{Confusion matrix of the predictions of the ResNet model (best-performing model) on BRACS test-set.}
    \label{cm}
\end{figure}

When we investigated different sampling methods like weight-batch samplers and batch-balanced samplers and their impact on classification performance. They had almost yielded  the same results with a 0.05 magnitude of difference. However, without them, performance degrades with an order of 10\% or more in the f1-score. As seen in Figure \ref{cm}, the ResNet50 model shows good performance when classifying N, FEA, and IC classes. However when making classification predictions, commonly, classes are mistaken for a different class entirely e,g ADH is mistaken for UDH or DCIS classes and DCIS is incorrectly predicted as ADH class.

In table \ref{trails_high_resulution_image}, the performance of different classification models using a high-resolution dataset is reported on the test set provided in the BRCAS dataset. The ConveNextTiny v2 model was able to achieve high performance with a f1-score of 61. Pyramid networks did show promising results compared to tiling methods, order of 10\% f1-score difference. This may lead to more investigation into pyramid models and different architectures.

In table \ref{sota_compare}, we compare results on the BRACS test dataset against state-of-the-art models such as ScoreNet and HACT-Net. Our ResNet50 model is fine-tuned on low resolution and shows good performance compared to SOTA models, with a magnitude of improvement of 10\% in the f1-score. This highlights that another way to handle a gigapixel dataset problem is to use appropriate resizing and remove pixels thus creating lower resolution images, and to take advantage of pre-trained models which are using smaller image sizes to enable effective transfer learning.

Tables \ref{upsample_1000_custom_split} and \ref{upsample_2000_custom_split} show the results of different models using a custom dataset split. In this custom split, we have combined the train-validation-test images as given in the BRACS dataset and then randomly assigned each image into a data subset using a 0.9,0.07,0.03 probability split for training, validation, and test sets respectively. For images assigned to the training subset, we performed multiple experiments where each class was upsampled to 1000 or 2000 samples per class. During upsampling, EffecientNet was able to show good performance during both experiments. With 1000 samples per class, the  model  demonstrated a 0.77 f1 score. Additionally, with 2000 samples per class, the model showed even greater performance with a 0.96 f1 score. This indicated that upsampling samples per image class can contribute significantly to better classification performance.

\begin{table}[h!]
    \centering
    \caption{Test results for the original BRACS dataset with the state-of-the-art models.}
    \label{sota_compare}

    \begin{tabular}{c|c}
        \toprule
        \textbf{Model} & \textbf{F1-score} \\
        \midrule
        \textbf{ResNet50 (Ours)} & \textbf{0.65}  \\
        ScoreNet \cite{stegmuller2023scorenet} & 0.64 \\
        HACT-Net \cite{pati2020hact} & 0.62 \\
        \bottomrule
    \end{tabular}
\end{table}

\begin{table}[h!]
    \centering
    \caption{Test results for the original BRACS dataset split using high-resolution images.}
    \label{trails_high_resulution_image}

    \begin{tabular}{p{2cm}cp{2cm}p{1cm}cp{1.2cm}p{1cm}}
        \toprule
        \textbf{Model} & \textbf{Image Size } & \textbf{Special Augmentation/ Preprocessing}  \textbf{Dropout rate} & \textbf{Loss Function} & \textbf{Accuracy} & \textbf{F1-score} \\
        \midrule
        ResNet50 Pyramid & (512,512) & CUTMIX and MIXUP & 0.45 & LabelSmoothing & 0.59 & 0.59 \\
        \textbf{ConvNextTiny V2} & \textbf{(512,512)} & \textbf{NO} & \textbf{0.45} & \textbf{Focal Loss} & \textbf{0.61} & \textbf{0.6} \\
        ResNet50 & (512,512) & NO & 0.45 & Focal Loss & 0.57 & 0.57 \\
        ResNet50 with Tiling & (1024,1024) & NO & 0.45 & Focal Loss & 0.5 & 0.49 \\
        \bottomrule
    \end{tabular}
\end{table}

\begin{table}[h!]
    \centering
    \caption{Test results for the custom BRACS dataset split with 1000 upsampling.}
    \label{upsample_1000_custom_split}
    \begin{tabular}{c|c|c|c}
        \toprule
        \textbf{Model} & \textbf{Accuracy} & \textbf{F1-score} & \textbf{Sensitivity} \\
        \midrule
        Xception & 0.716 & 0.713 & 0.712 \\
        \textbf{EfficientNet} & \textbf{0.768} & \textbf{0.77} & \textbf{0.77} \\
        ResNet50 & {0.73} & {0.725} & 0.729 \\
        InceptionResNet & 0.725 & 0.725 & 0.724 \\
        \bottomrule
    \end{tabular}
\end{table}

\begin{table}[h!]
    \centering
    \caption{Test results for the custom BRACS dataset split with 2000 upsampling.}
    \label{upsample_2000_custom_split}
    \begin{tabular}{c|c|c|c}
        \toprule
        \textbf{Model} & \textbf{Accuracy} & \textbf{F1-score} & \textbf{Sensitivity} \\
        \midrule
        Xception & 0.884 & 0.884 & 0.887 \\
        EfficientNet & 0.955 & 0.955 & 0.956 \\
        \textbf{ResNet50} & \textbf{0.962} & \textbf{0.962} & \textbf{0.962} \\
        InceptionResNet & 0.957 & 0.957 & 0.958 \\
        \bottomrule
    \end{tabular}
\end{table}

\begin{table}[h!]
    \centering
    \caption{Test results for the original BRACS dataset split using low-resolution images.}
    \label{trails_low_resulution_image}
    \begin{tabular}{p{2cm}cp{2cm}p{1cm}cp{1.2cm}p{1cm}}
    \toprule
    \textbf{Model} & \textbf{Image Size} & \textbf{Special Augmentation/ Preprocessing} & \textbf{Dropout rate} & \textbf{Loss Function} & \textbf{Accuracy} & \textbf{F1-score} \\
    \midrule
    EfficientNet & (1024,512) & NO & 0.35 & Focal loss & 0.59 & 0.59 \\
    EfficientNet & (512,512) & NO & 0.45 & Focal loss & 0.62 & 0.62 \\
    EfficientNet & (512,512) & NO & 0.5 & Focal loss & 0.51 & {0.49} \\
    ResNet50 & (512,512) & NO  & 0.45 & CrossEntropy & 0.57 & 0.58 \\
    \textbf{ResNet50} & \textbf{(512,512)} & \textbf{NO} & \textbf{0.45} & \textbf{Focal Loss} & \textbf{0.66} & \textbf{0.65} \\
    ResNet50 & (512,512) & CUTMIX and MIXUP & 0.45 & LabelSmoothing & 0.63 & 0.62 \\
    ResNet50 & (512,512) & Stain Normalization & 0.45 & CrossEntropy & 0.59 & 0.57 \\
    ResNet50 Pyramid  & (512,512) & CUTMIX and MIXUP & 0.45 & LabelSmoothing & 0.63 & 0.63 \\
    ResNet50 with Tiling & (1024,1024)  & NO & 0.45 & Focal Loss & 0.59 & 0.47 \\
    InceptionResNet & (512,512) & NO & 0.45 & Focal Loss & 0.56 & 0.54 \\
    \bottomrule
    \end{tabular}
\end{table}

\section{Challenges and Limitations}
 Different laboratories and institutions use different staining and imaging protocols for H\&E stained histology images. Furthermore, tissue preparation, staining duration, reagents, and imaging equipment can affect image appearance and quality, making it hard to establish consistent patterns or features. Another issue is that pathologists or domain experts must manually label and classify histology images, which can lead to inter-observer variability and subjectivity, or even annotation inconsistencies between different reviewers. Also, different datasets may have quality assurance processes or labelling criteria, making it harder to establish consistent ground truth class labels across different image datasets. These obstacles prevent us from generalising our study across different datasets. However, we are excited to experiment on specific classes like Normal (N), Pathological Benign (PB), Usual Ductal Hyperplasia (UDH), Ductal Carcinoma in Situ (DCIS), and Invasive Carcinoma (IC) from the previous datasets.

\section{Conclusion}
During this research project, we aimed to answer one main research question:
``Can tuned Xception, EfficientNet, ResNet50, ConvNextTiny, and Inception-ResNet models achieve comparable performance to current state-of-the-art multi-instance learning (MIL) approaches for the application of classifying  breast cancer histology gigapixel images?" The research investigation that was undertaken showed that these models are suitable for the task of classifying images of breast cancer histology slides.

We performed three different approaches while undertaking this research. For the first approach, we experimented with the default BRACS dataset splits, and in our test results the ResNet50 model outperformed SOTA models achieving 65\% F1-score. For our second approach, we split the dataset into custom train-val-test splits and also upsampled the dataset to 1000 samples per class, and in our test results the EfficientNet showed the best classification performance, achieving 77\% F1-score and 77\% sensitivity. For our third approach, we randomly split the dataset into train-val-test custom splits (90\%-7\%-3\%) and also upsampled the dataset to 2000 samples per class. In our test results using this data split, the RestNet50 model gave the best classification performance, achieving 96.2\% F1-score and 96.2\% sensitivity.

We have demonstrated that normal patch-based model training processes with the inclusion of transfer learning are capable of achieving high classification results when applied to gigapixel breast cancer histology images.

\section{Conflict of Interest}

The authors do not have any personal or financial interests related to the subject matters discussed in this manuscript.

\section{Data Availability}

The data that support the findings of this study are openly available in ``ICNAR-CNR'' at https://www.bracs.icar.cnr.it/ and http://doi.org/10.48550/arXiv.2111.04740, reference number \cite{brancati2022bracs}.

\bibliographystyle{plain}



\end{document}